\documentclass[conference]{IEEEtran}
\usepackage{color}
\usepackage[utf8]{inputenc} 
\usepackage[T1]{fontenc}    
\usepackage{hyperref}       
\usepackage{url}            
\usepackage{booktabs}       
\usepackage{amsfonts}       
\usepackage{nicefrac}       
\usepackage{microtype}      

\usepackage{graphicx}

\usepackage{amsmath}
\usepackage{bbm}
\usepackage[colorinlistoftodos]{todonotes}
\usepackage{comment}
\usepackage{epsfig} 
\usepackage{amsmath} 
\usepackage{amssymb}  
\usepackage{multicol}
\usepackage{multirow}
\usepackage{color}
\usepackage{mathrsfs}
\usepackage{xparse}
\usepackage{bm}
\usepackage{algorithm}
\usepackage{algorithmicx}
\usepackage{algpseudocode}
\usepackage[font=small,labelfont=bf,tableposition=top]{caption}
\usepackage[font=footnotesize]{subcaption}

\DeclareMathOperator{\E}{\mathbb{E}}

\usepackage{mathtools}

\usepackage[colorinlistoftodos]{todonotes}
\usepackage{optidef}
\usepackage{gensymb}

\usepackage{etoolbox}
\newtoggle{comments}
\settoggle{comments}{true} 
\iftoggle{comments}{
  \newcommand {\alberto}[1]{{\color{orange}{~Alberto: #1}\normalfont}}
  \newcommand {\bjin}[1]{{\color{blue}{~Baihong: #1}\normalfont}}
  \newcommand {\dan}[1]{{\color{olive}{~Dan: #1}\normalfont}}
  \newcommand {\yuxin}[1]{{\color{violet}{~Yuxin: #1}\normalfont}}
  \newcommand {\poolla}[1]{{\color{green}{~Kameshwar: #1}\normalfont}}
  \newcommand {\red}[1]{{\color{red}{#1}\normalfont}}
}{
  \newcommand {\alberto}[1]{{}}
  \newcommand {\bjin}[1]{{}}
  \newcommand {\dan}[1]{{}}
  \newcommand {\yuxin}[1]{{}}
  \newcommand {\poolla}[1]{{}}
  \newcommand {\red}[1]{{}}
}

\usepackage{acronym}
\acrodef{HVAC}{Heating, Ventilation and Air Condtioning}
\acrodef{SVM}{Support Vector Machine}
\acrodef{AE}{AutoEncoder}
\acrodef{RNN}{Recurrent Neural Network}
\acrodef{RUL}{Remaining Useful Life}
\acrodef{OC-SVM}{One-class Support Vector Machine}
\acrodef{CPS}{Cyber-Physical System}
\acrodef{DE}{Differential Evolution}
\acrodef{RBF}{Radial Basis Function}
\acrodef{PdM}{Predictive Maintenance}
\acrodef{FDD}{Fault Detection and Diagnosis}
\acrodef{LDA}{Linear Discriminant Analysis}
\acrodef{PCA}{Principal Component Analysis}
\acrodef{SPC}{Statistical Process Control}
\acrodef{AHU}{Air Handling Unit}

\ifCLASSINFOpdf
\else
\fi
\begin{document}
%


\title{\LARGE \bf
Detecting and Diagnosing Incipient Building Faults Using Uncertainty Information from Deep Neural Networks
}


\author{Baihong~Jin$^1$~~Dan~Li$^2$~~Seshadhri~Srinivasan$^3$~~{See-Kiong~Ng}$^2$\\{Kameshwar~Poolla}$^{1,3}$~~{Alberto~Sangiovanni-Vincentelli}$^{1,3}$\\
$^{1}$Department of EECS, University of California, Berkeley\\
$^{2}$Institute of Data Science, National University of Singapore\\
$^{3}$The Berkeley Education Alliance for Research in Singapore\\
}


\IEEEoverridecommandlockouts
\IEEEpubid{\makebox[\columnwidth]{978-1-5386-8357-6/19/\$31.00~\copyright2019 IEEE \hfill} \hspace{\columnsep}\makebox[\columnwidth]{ }}
\maketitle
\IEEEpubidadjcol
\begin{abstract}
Early detection of incipient faults is of vital importance to reducing maintenance costs, saving energy, and enhancing occupant comfort in buildings. Popular supervised learning models such as deep neural networks are considered promising due to their ability to directly learn from labeled fault data; however, it is known that the performance of supervised learning approaches highly relies on the availability and quality of labeled training data. In \ac{FDD} applications, the lack of labeled incipient fault data has posed a major challenge to applying these supervised learning techniques to commercial buildings. To overcome this challenge, this paper proposes using Monte Carlo dropout (MC-dropout) to enhance the supervised learning pipeline, so that the resulting neural network is able to detect and diagnose unseen incipient fault examples. We also examine the proposed MC-dropout method on the RP-1043 dataset to demonstrate its effectiveness in indicating the most likely incipient fault types. 
\end{abstract}

%
\IEEEpeerreviewmaketitle
\section{Introduction}\label{sec:introduction}

Building faults whose impact are less perceivable and/or hinder regular operations are called \textit{soft faults}~\cite{Comstock2001,YanKe2014ARX}. These soft faults, especially in their incipient phase, are hard to detect as their signatures are not generally obvious (due to their magnitudes) and are lurking under measurement/system noise or feedback control actions~\cite{he2018incipient,watanabe1989incipient}. Nevertheless, they will impact energy consumption, system performance, and maintenance costs adversely in the long-run if left undetected and unattended~\cite{jia2018design}. In addition, they can lead to costly maintenance and undesirable replacement operations. Therefore, it is an important and challenging task to design methods to detect and diagnose incipient soft faults during their incipient stage for various building systems, such as chillers and \acp{AHU}.

\acf{FDD} methods in the literature can be broadly classified into three categories: {\em{(i)}} model-based, {\em{(ii)}} signal-based, and {\em{(iii)}} data-driven~\cite{Dai2013from,Gao2015asurvey}. Model-based methods depend on explicit physical models at the device levels and use correlation tests on the input-ouput data to detect faults~\cite{Qin2005afault,Saththasivam2008predictive,Ru2008variable}. While fault-diagnosis can also be performed with the model used for detection, developing detailed models is a time-consuming and daunting process, especially for complex \acp{CPS} like buildings. Authors in~\cite{Dumidu2014mining} point out that model-based methods are not as practical as data-driven methods in terms of applying the \ac{FDD} techniques to real buildings. Signal-based \ac{FDD} methods find sensor measurement signatures to indicate faults. Signal-based \ac{FDD} combining wavelet transformation and principal component analysis was presented in~\cite{Shun2014amodel}. Although the methods achieved good performance, extracting relevant signatures and signals that indicate faulty condition is a daunting task for complex systems such as buildings.
In data-driven \ac{FDD} approaches, when labeled fault data are available, a \ac{FDD} task are usually modeled as a multiclass classification problem. Then a supervised learning method can be employed to learn a classifier to recogize the faults. Many supervised methods such as multivariate regression models \cite{Mustafaraj:2010}, Bayes classifiers~\cite{Hill:2007,Zhao:2013,Xiao:2014}, neural networks (NN)~\cite{Fan:2010,Zhu:2012,Du:2014}, Fisher Discriminant Analysis (FDA)~\cite{Du:2008}, Gaussion Mixture Models~\cite{Jaikumar:2011}, Support Vector Data Description (SVDD)~\cite{Yang:2013,Zhao:2014}, and Support Vector Machines (SVM)~\cite{Liang:2007,Han:2010,Chen:2011,Yan:2014,Mulumba:2015} have been proposed to classify the faults. Recently, Li~et~al.~\cite{li2016fault} proposed a tree-structured learning method that not only recognizes faults but also their severity levels; however, it is hard in practice to obtain such a well-labeled dataset that include incipient faults. Researchers have also unsupervised approaches using \ac{PCA}, \ac{SPC}, and auto-encoders for \ac{FDD}. Depending only on positive (healthy) class data, such methods have found their use in detecting anomalies; however, they still lack the ability to diagnose these anomalies.

A review of the literature reveals that data-driven approaches relying on supervised learning are promising methods due to their ability to classify and differentiate data with multiple labels. However, in order to train a well-performing model, large amount of labeled data is typically needed, which is not always easy to obtain. Furthermore, although supervised learning tends to perform well on known (in-distribution) data patterns, the unseen (out-of-distribution) data may lead to unexpected prediction behaviors. In the context of \ac{FDD}, an incipient fault example not seen in the training phase may fool the classifier into wrong belief, which is certainly not desirable for \ac{FDD} applications. Although this problem can be conceptually alleviated by using a larger, more comprehensive training dataset, in practice it is technically infeasible to obtain fault data of all different fault types, and of all possible severity levels, especially for complex building systems such as chillers.

The aforementioned reasons motivate us to devise a method that can make full use of the available training data. The resulting classifier should not only be good that classifying in-distribution data points, but is also able to give reasonable diagnostic suggestions as well as its prediction uncertainty for out-of-distribution fault examples. The contribution of this paper is two-fold: 
\begin{itemize}
    \item We propose using the uncertainty information given by machine learning models to detect and diagnose unknown incipient faults in building systems.
    \item To effectively estimate the uncertainty information for this purpose, we propose using MC-dropout networks. The approach requires few modifications to the standard deep learning pipeline, making it attractive for real-world \ac{FDD} applications. A case study is conducted on the ASHRAE~RP-1043 dataset, which demonstrates the effectiveness of our approach.
\end{itemize}

The rest of the paper is organized as follows. In Sec.~\ref{sec:approach}, we give the necessary background about the MC-d.ropout approach and discuss how it can help us identify faults under development. Next, we describe and analyze the RP-1043 dataset in Sec.~\ref{sec:rp1043}, and then present a case study in Sec.~\ref{sec:experiment} with extensive experiment results to show the effectiveness our proposed approach behaves on the RP-1043 dataset. We later conclude the paper in Sec.~\ref{sec:conclusion} and also discuss potential future steps.

\section{Monte Carlo Dropout Approach}\label{sec:approach}
\subsection{Neural Network Classifiers}
\begin{figure}[tb]
    \centering
    \includegraphics[width=0.9\linewidth]{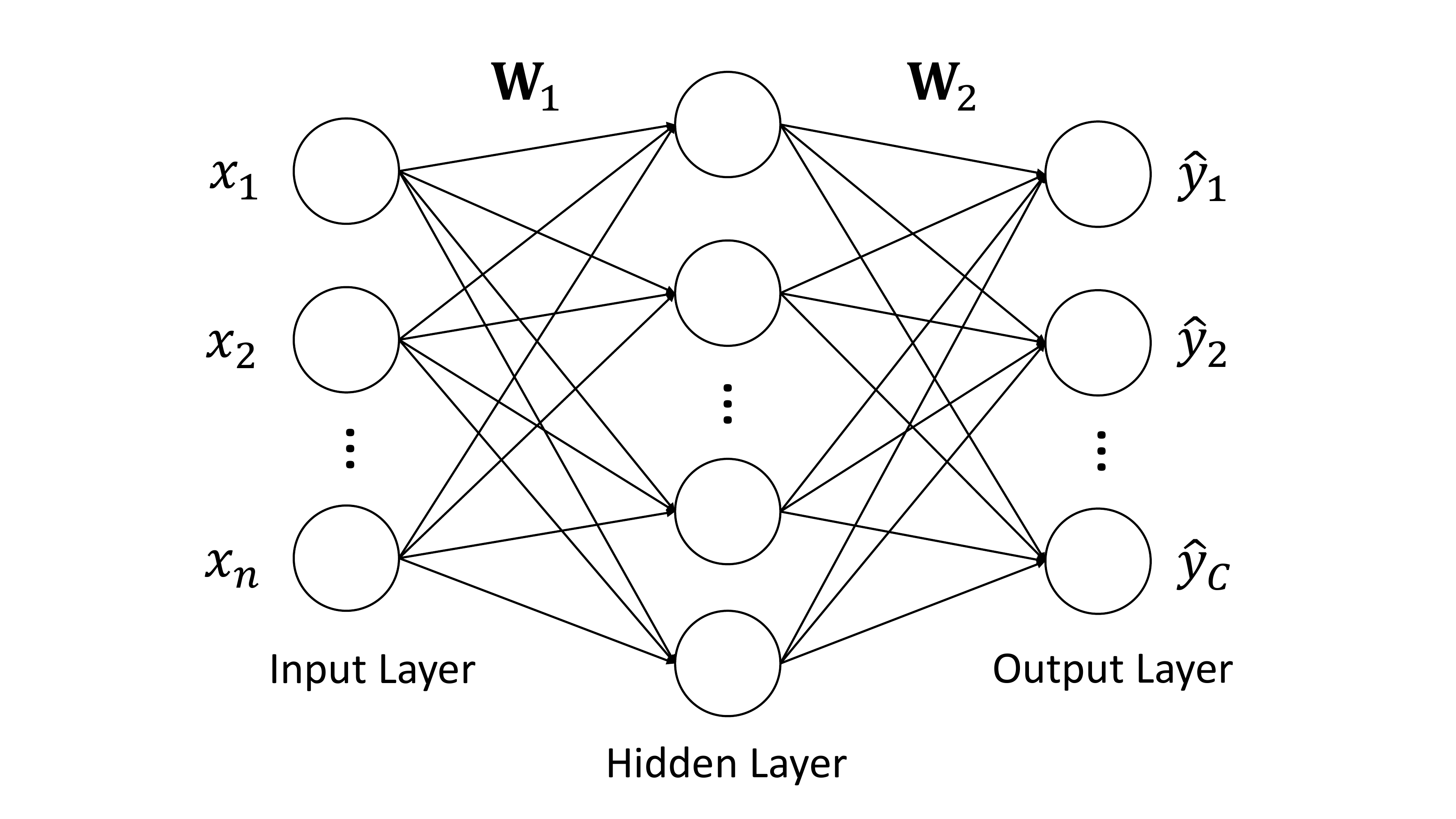}
    \caption{An example of a simple feedforward neural network with one hidden layer. The intercept parameter $b_1$ associated with the hidden layer is omitted for brevity.}
    \label{fig:nn}
\end{figure}

The MC-dropout network to be introduced later is based on the classic feed-forward neural network model, a.k.a.~\textit{multilevel perceptron}. A network usually consists of two or more layers, and can be described as a series of functional transformations on the input vector. We take a simple feedforward neural network with one hidden layer, shown in Fig.~\ref{fig:nn}, as an example. The value of the hidden layer $\bm{h}$ is computed from the input vector $\bm{x}$ in the following way.
\begin{align}
    \bm{h} = g(\mathbf{W}_1\bm{x}+b_1),
\end{align}
where $\mathbf{W}_1$ defines an affine transformation to $\bm{x}$, and $g$ is an \textit{activation function} that is typically nonlinear and applied element-wise $\mathbf{W}_1\bm{x} + b_1$. In modern neural networks, the rectified linear unit or ReLU defined by $g(t) = \max\{0,t\}$ is usually used as the activation function~\cite{Goodfellow-et-al-2016}.

For a multiclass classification problem with $C$ classes, we need the neural network classifier to output a vector $\bm{\hat{y}} = (\hat{y}_1, \hat{y}_2,\ldots,\hat{y}_C)$ representing a discrete probability distribution. We require that each $\hat{y}_i=P(y=i\,\vert\,\bm{x})\in(0,1)$, and $\sum_{i=1}^C\hat{y}_i=1$, i.e.~these probabilities sum up to $1$. A softmax activation function is usually used at the last layer to obtain the desired $\bm{\hat{y}}$. Let $\bm{z}=\mathbf{W}_2\bm{h}$ be the \textit{activation} of the last layer\footnote{The bias coefficient is not needed for this layer because adding a bias to every element of $z$ will not change the softmax output.}. Under the softmax transformation, for each class $i=1,2,\ldots,C$, $z_i = \log P(y=i\,\vert\,\bm{x})$. And then the softmax probability for class $i$ is given by
\begin{align}
    P(y=i\,\vert\,\bm{x}) = \frac{\exp(z_i)}{\sum_{j=0}^C\exp(z_j)}.
\end{align}
To train such models for multiclass classification tasks, we usually minimize the cross-entropy loss as below
\begin{align}
    L = -\frac{1}{N}\sum_{j=1}^N\sum_{i=1}^Cy_i^{(k)}\log y_i^{(k)},
\end{align}
in order to maximize the log-likelihood of the softmax distribution, over the training samples.

\begin{figure}[tb]
    \centering
    \begin{subfigure}[t]{0.48\linewidth}
    \centering
    \includegraphics[width=\linewidth]{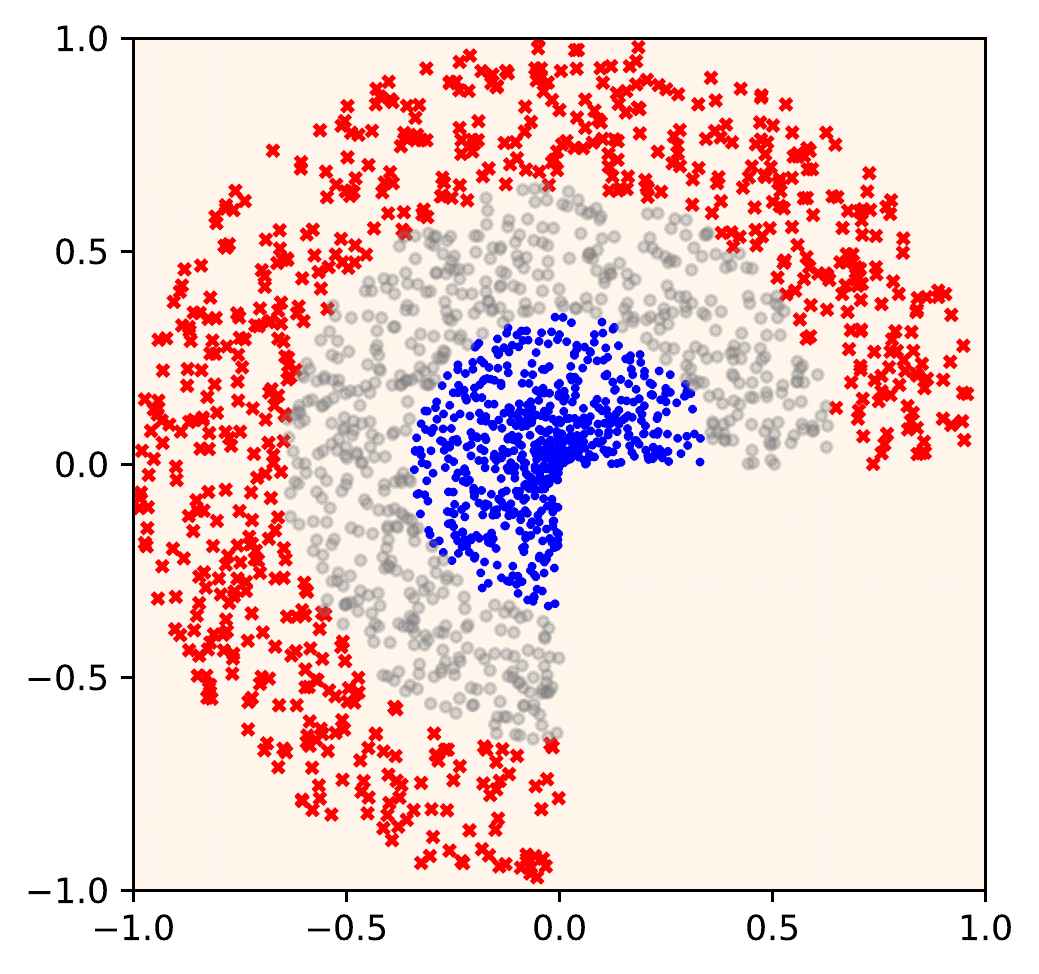}
    \caption{Distribution of data points}\label{fig:toy-example-distribution}
    \end{subfigure}
    \begin{subfigure}[t]{0.48\linewidth}
    \centering
    \includegraphics[width=\linewidth]{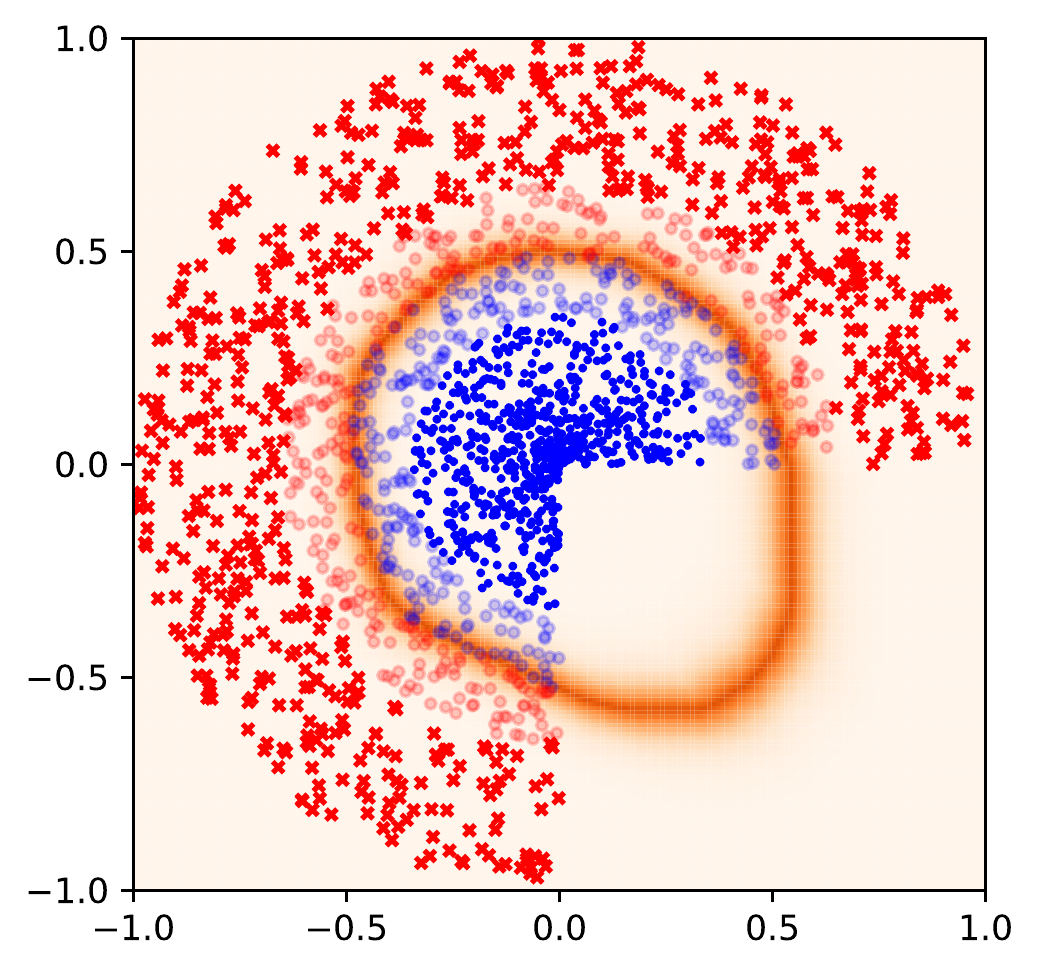}
    \caption{Decision boundary}\label{fig:toy-example-boundary}
    \end{subfigure}
    \caption{A toy example in 2 dimensions. (a) The healthy state data (in blue) are confined in a circle $r=0.3$ centered at the origin. The severe fault data (in red) reside outside $r=0.7$. In between the blue and the red points are the intermediate states (gray) that are not observed in the training distribution. (b) The decision boundary (orange) given by a trained non-dropout neural network classifier. Intermediate-level fault data points classified as healthy are shown in light blue, and those classified as faulty by the decision boundary are shown in light red.}\label{fig:toy-nn}
\end{figure}

Here, we illustrate the overconfidence problem of neural networks by using a toy example in two dimensions shown in Fig.~\ref{fig:toy-example-distribution}. 
As displayed in the plot, the decisions boundary forms a narrow band running across the intermediate states, separating the healthy and severe fault examples; most of the gray points that are not on the decision boundary are either classified as healthy or faulty with high confidence (with network output very close to $0$ or $1$). By using this decision boundary, the gray points that are closer to the origin will be classified as healthy, while most others will be reckoned faulty; see Fig.~\ref{fig:toy-example-boundary} Only the few that reside on the boundary will be considered ambiguous because the fault probabilities are close to $0.5$. Although this classification model can recognize some faulty conditions, it is not enough for detecting incipient faults.
In addition, the trained model also shows high confidence on the bottom right region where no data are available. It seems that the network has extrapolated its learned pattern to the unseen region. Such extrapolation could be sometimes dangerous---the data distribution might be totally different in that region. It is more desirable for the model to be cautious about what the available data cannot offer.

\subsection{Estimating Predictive Uncertainty}

Incipient faults are often characterized by small deviations from fault-free conditions. As a result, their behaviors often resemble both fault-free and fault patterns, which could confuse a classifier.
To detect incipient faults, it would be desirable to have a neural network model that can tell its uncertainty in the predictions; however, as previously mentioned, neural networks are usually poor at telling the predictive uncertainties, and often tend to be overly confident in its predictions. 
The standard and state-of-the-art approach for estimating predictive uncertainty is to use Bayesian neural networks, whose goal is to learn a distribution over weights; however such approaches are typically computationally expensive compared to standard (non-Bayesian) neural neural networks, and do not naturally fit into the standard training pipeline of today's deep learning frameworks, which could limit their use. Details of Bayesian neural networks are beyond the scope of this paper, and we refer interested readers to works~\cite{lakshminarayanan2017simple} and references therein for more in-depth discussions.

Besides Bayesian neural networks, there are also approaches that use ensembles of models for estimating predictive uncertainty~\cite{lakshminarayanan2017simple}. The MC-dropout approach to be described next also belongs to these ensemble methods. Such ensemble approaches typically involves some randomization, either in the base learners themselves or in the data used to train each base learner. The idea behind is simple and intuitive: use an ensemble of individual models to obtain multiple predictions, and use the empirical variance of predictions as an approximate measure of uncertainty~\cite{lakshminarayanan2017simple}. 
For these ensemble methods to work, the individual classifiers must exhibit \textit{diversity} among themselves; the diversity allows individual classifiers to generate different decision boundaries.
Due to the randomness of decision boundaries learned by each individual classifier, the ensemble can hopefully give a high predictive uncertainty on out-of-distribution data points, which provides us with a way to detect them. Ensemble learning requires a combination of many diverse base learners in order to build an ensemble classifier. As a result, the idea of ensemble learning is typically adopted in learning schemes where the base learners can be fitted quickly, e.g.,~random forests, which seems to limit the use of neural networks in an ensemble model because they are time-consuming to train.

\subsection{The MC-dropout Approach}

\begin{figure}[tb]
    \centering
    \includegraphics[width=0.9\linewidth]{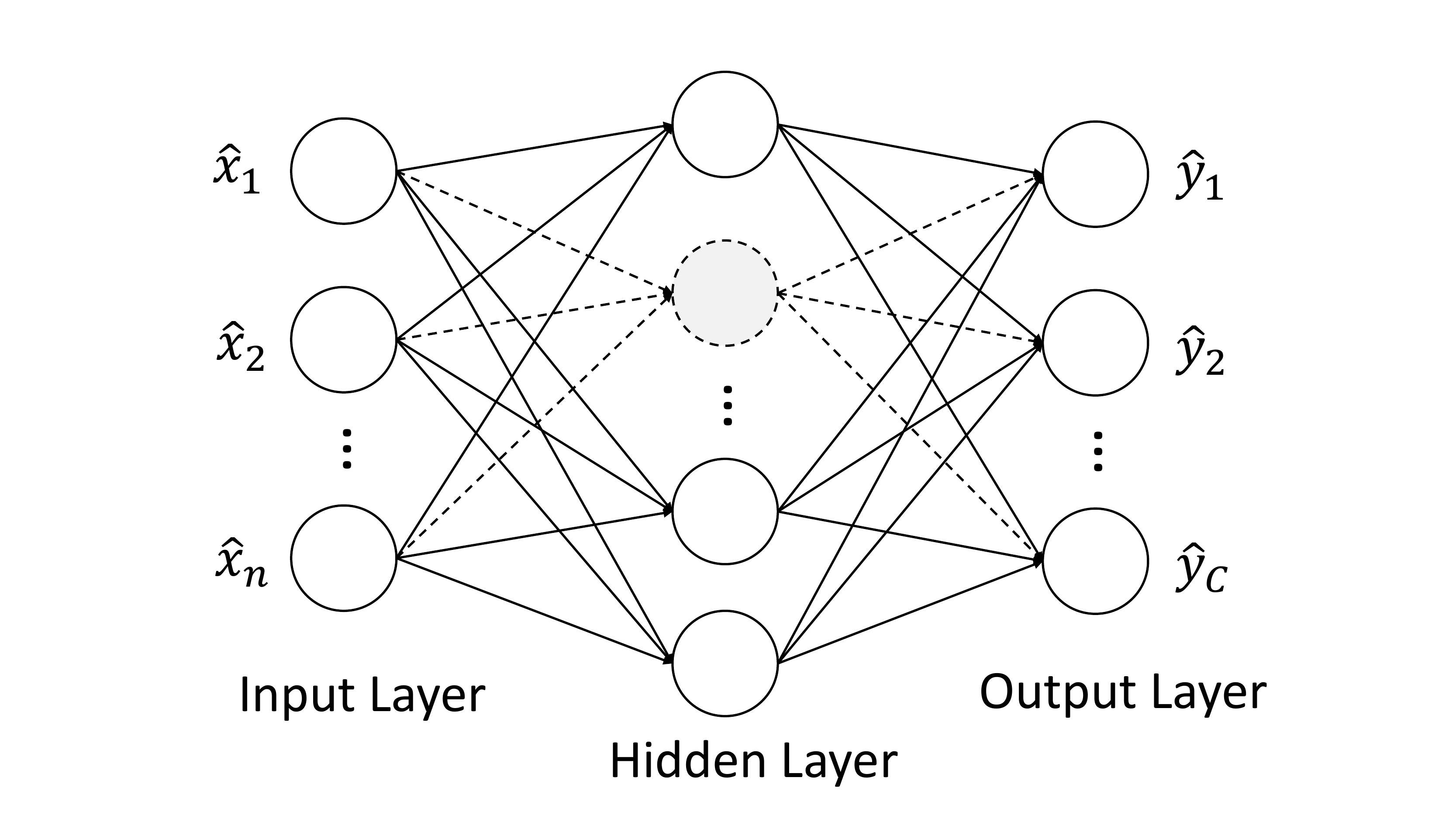}
    \caption{A simple feedforward neural network with dropout.}
    \label{fig:dp-nn}
\end{figure}

Dropout~\cite{srivastava2014dropout} is a popular and powerful regularization technique to prevent over-fitting neural network parameters. The key idea is to randomly drop units along with their connections from the network during training; see Fig.~\ref{fig:dp-nn} for an illustration. Each individual hidden node is dropped at a probability of $p$, i.e.~the dropout rate. The training and inference procedure is then run as usual. In effect, the dropout technique provides an inexpensive approximation to training and evaluating an ensemble of exponentially many neural networks.

The dropout mechanism offers a way to incorporate intrinsic randomization into neural network models; recently, Gal and Ghahramani proposed using MC-dropout~\cite{gal2016uncertainty} to estimate a network's predictive uncertainty by using dropout at test time. During testing, we treat a model $\mathcal{M}$ trained using dropout as if we were using it during the training phase. Each time we forward pass a given input $\bm{x}$ through the network, each hidden node in the network will be dropped at a probablity of $p$ and we will obtain a random output $\hat{\bm{y}}$. By repeating the same process for $T$ times, we will obtain $T$ i.i.d.~sampled output vectors $\hat{\bm{y}}^{(1)},\hat{\bm{y}}^{(2)},\ldots,\hat{\bm{y}}^{(T)}$. Their predictive mean $\E[\hat{\bm{y}}] = \frac{1}{T}\sum_{k=1}^T\hat{\bm{y}}^{(k)}$ can be understood as the expected output given input $\bm{x}$, and the predictive variance $\frac{1}{T}\sum_{k=1}^T\left(\hat{\bm{y}}^{(k)}-\E[\hat{\bm{y}}]\right)$ can be used to measure the \textit{confidence} of $\mathcal{M}$ in its prediction. The larger the predictive variance is, the more uncertain is the network about its prediction.

\begin{figure}[tb]
    \centering
    \includegraphics[width=0.9\linewidth]{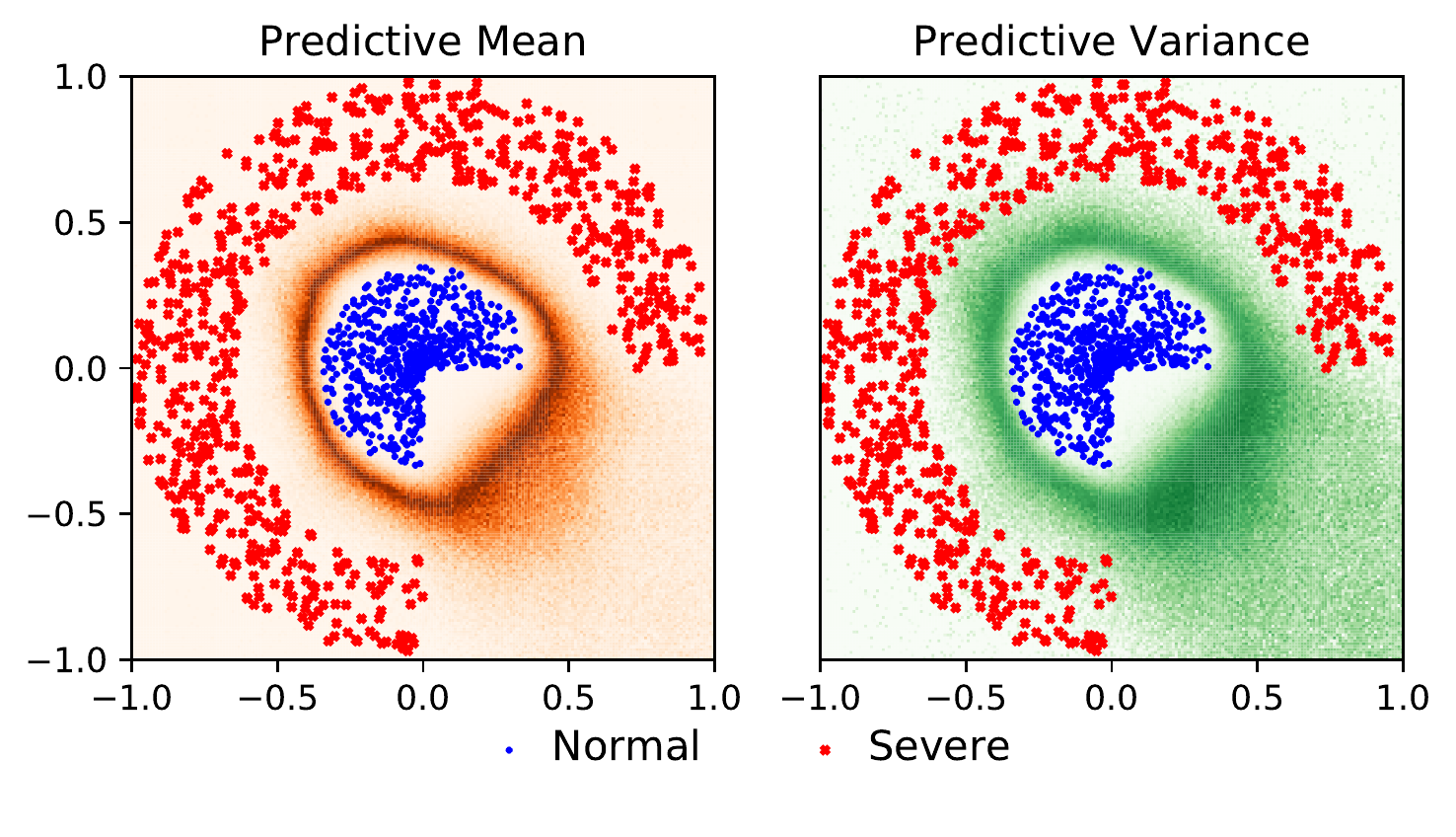}
    \caption{The spatial distribution of the predictive mean (in orange shades) and the predictive variance (in green shades) of an MC-dropout model trained on the aforementioned toy example data. For the predictive mean plot, the color intensity signifies the proximity between the predictive mean at a point and $0.5$. In the predictive variance plot, the more intense the color, the higher predictive variance at a given point. }\label{fig:toy-nn-mc}
\end{figure}

To illustrate the approach, we trained a simple feed-forward neural network with MC-dropout on the previously introduced 2D toy example; the results are shown in Fig.~\ref{fig:toy-nn-mc}. 
It can be observed that the decision boundary in orange (given by points with their predictive mean close to $0.5$) looks similar to the boundary given by the non-dropout network (see Fig.~\ref{fig:toy-nn}). Here the predictive uncertainty information will play a crucial role in suggesting potential faults. We can see from the right panel of Fig.~\ref{fig:toy-nn-mc} that the regions where intermediate states reside, especially in the vicinity of the healthy region, are associated with elevated predictive variance (shown as green shades). In addition, high predictive variance is present in the bottom right region, where no data points have been observed. These behaviors indicate that the trained MC-dropout network is suspicious about, while still being cautious, potential faults in such regions, although not much prior information is provided by the data distribution for training the network model. Such suspicion is important and can serve as alarms for potential faults, and further inspection and maintenance measures shall be taken if necessary.

The example above demonstrates that MC-dropout technique can provide hints about incipient faults, which is regular neural networks fall short of. But just detecting faults or anomalies is not our sole purpose, nor does it make the MC-dropout approach distinctive from other approaches. Classic anomaly detection methods, such as \ac{PCA} and autoencoders, can also suggest out-of-distribution inputs; however, these approaches do not possess discriminative ability between fault conditions. Next, we will demonstrate through a case study on the ASHRAE~RP-1043 Dataset that the MC-dropout approach can not only detect, but also provide preliminary diagnosis about, underlying health problems of an industrial chiller system.

\begin{table}[tb]
\caption{Descriptions of the variables used as features}
\label{tbl:features}
\begin{tabular}{lll}
\hline
Sensor & Description & Unit \\ \hline
TEI & Temperature of entering evaporator water & \degree F \\
TEO & \begin{tabular}[c]{@{}l@{}}Temperature of leaving evaporator\\   water\end{tabular} & \degree F \\
TCI & \begin{tabular}[c]{@{}l@{}}Temperature of entering condenser\\   water\end{tabular} & \degree F \\
TCO & \begin{tabular}[c]{@{}l@{}}Temperature of leaving condenser\\   water\end{tabular} & \degree F \\
Cond Tons & \begin{tabular}[c]{@{}l@{}}Calculated Condenser Heat\\   Rejection Rate\end{tabular} & Tons \\
Cooling Tons & Calculated City Water Cooling Rate & Tons \\
kW & Compressor motor power consumption & kW \\
FWC & Flow Rate of Condenser Water & gpm \\
FWE & Flow Rate of Evaporator Water & gpm \\
PRE & \begin{tabular}[c]{@{}l@{}}Pressure of refrigerant in\\   evaporator\end{tabular} & psig \\
PRC & \begin{tabular}[c]{@{}l@{}}Pressure of refrigerant in\\   condenser\end{tabular} & psig \\
TRC & sub Subcooling temperature & \degree F \\
T\_suc & Refrigerant suction temperature & \degree F \\
Tsh\_suc & \begin{tabular}[c]{@{}l@{}}Refrigerant suction superheat\\   temperature\end{tabular} & \degree F \\
TR\_dis & Refrigerant discharge temperature & \degree F \\
Tsh\_dis & \begin{tabular}[c]{@{}l@{}}Refrigerant discharge superheat\\   temperature\end{tabular} & \degree F \\ \hline
\end{tabular}
\end{table}
%

\section{Chiller System and Dataset}\label{sec:rp1043}

\begin{figure}[tb]
    \centering
    \includegraphics[width=\linewidth]{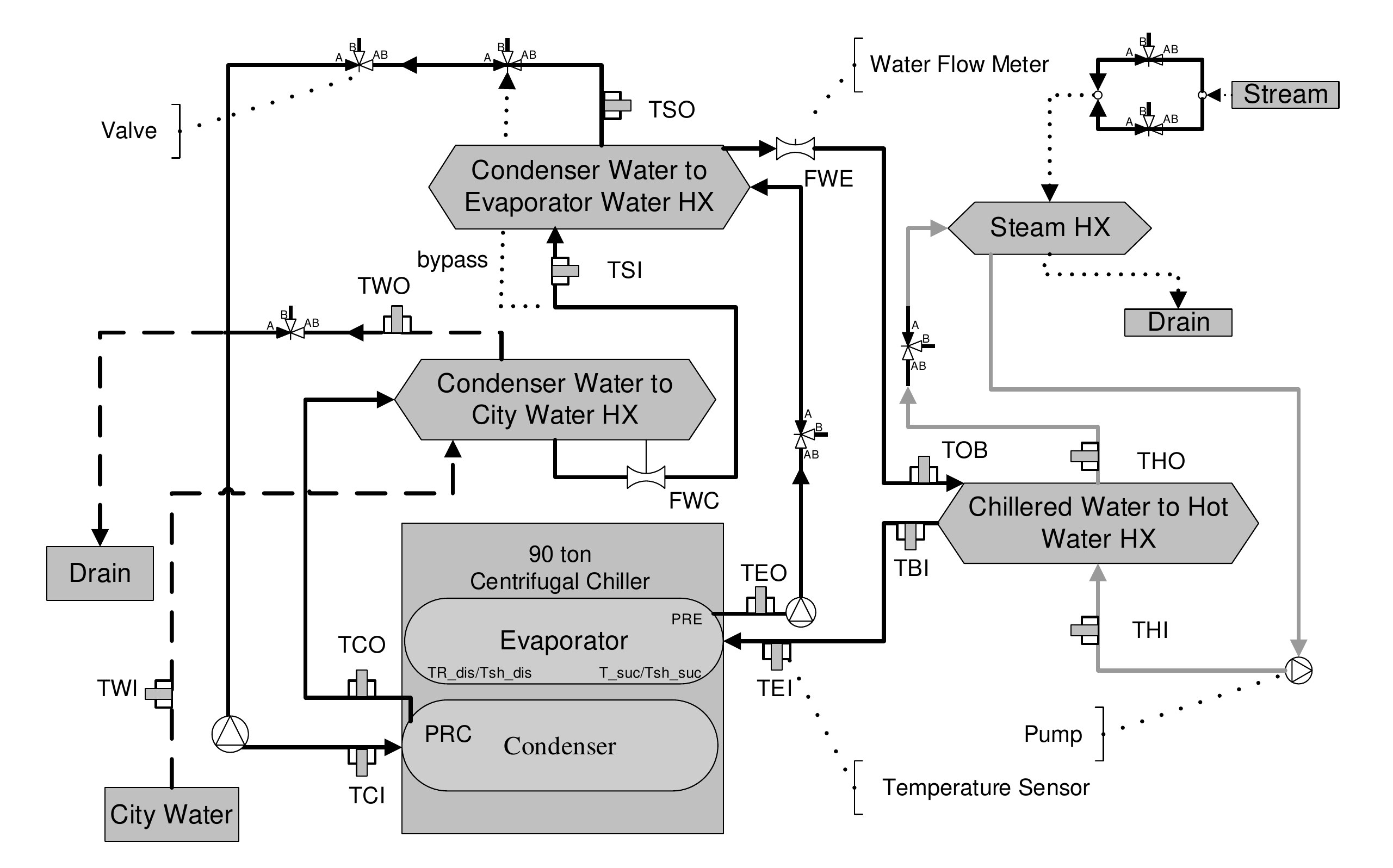}
    \caption{Schematic of the cooling system test facility and sensors mounted in the related water circuits.}
    \label{fig:chiller-sensor}
\end{figure}

\begin{figure*}[t]
    \centering
    \includegraphics[width=\textwidth]{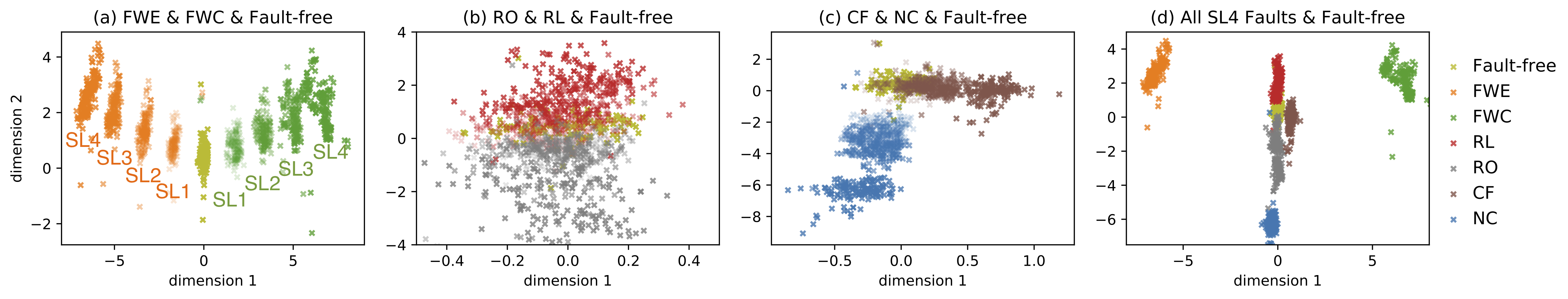}
    \caption{A visualization of the RP-1043 dataset, including the six faults being studied as well as the fault-free data. To visualize these high-dimensional data points, we employed \ac{LDA} for dimensionality reduction. The color intensity of a data point signifies the severity of the corresponding fault. In (a), the SL1-SL4 data form clusters that are easily distinguishable. In (b) and (c) faults of different severity levels are harder to separate.}
    \label{fig:LDA}
\end{figure*}

We used the ASHRAE~RP-1043 Dataset~\cite{comstock1999development} to test out the proposed MC-dropout approach for incipient fault detection.
In RP-1043, sensor measurements of a typical cooling system---a 90-ton centrifugal water-cooled chiller---were recorded under both fault-free and various fault conditions. The 90-ton chiller is representative of chillers used in larger installations~\cite{Comstock2002fault}, and consists of the following parts: evaporator, compressor, condenser, economizer, motor, pumps, fans, and distribution pipes etc. with multiple sensor mounted in the system. Fig.~\ref{fig:chiller-sensor} depicts the cooling system with sensors mounted in both evaporation and condensing circuits.

In RP-1043 experimental data, eight different types of process faults were injected into the chiller, and each fault was introduced at four levels of severity (SL1\,-\,SL4, from slightest to severest). In this study, we only included the six faults shown in Table~\ref{tbl:faults}, because an earlier study by Reddy~\cite{t2007development} found certain limitations with the excess oil and faulty TXV operation data. 

The condenser fouling (CF) fault was emulated by plugging tubes into condenser. The reduced condenser water flow rate (FWC) fault and reduced evaporator water flow rate (FWE) fault were emulated directly by reducing water flow rate in the condenser and evaporator. The refrigerant overcharge (RO) fault and refrigerant leakage (RL) fault were emulated by reducing or increasing the refrigerant charge respectively. The excess oil (EO) fault was emulated by charging more oil than nominal. And the non-condensable in refrigerant (NC) fault was emulated by adding Nitrogen to the refrigerant.

All faults were tested at 27 different operating conditions with varying chiller thermal load, chilled water outlet and inlet temperature settings. The data were collected at ten-second intervals, not only when the system has reached steady state, but also at transient states in between. We focused on only the steady-state data in this study. A more detailed review and discussion on this topic can be found in~\cite{t2007development}. The sixteen key features identified by RP-1043, as listed in Table~\ref{tbl:features}, were selected to train our neural network models.

\begin{table}[t]
\caption{The six chiller faults in our study}
\label{tbl:faults}
\begin{tabular}{lll}
\hline
 & \textbf{Fault} & \textbf{Normal Operation} \\ \hline
1 & Reduced Condenser Water Flow (FWC) & 270 gpm \\
2 & Reduced Evaporator Water Flow (FWE) & 216 gpm \\
3 & Refrigerant Leak (RL) & 300 lb \\
4 & Refrigerant Overcharge (RO) & 300 lb \\
5 & Condenser Fouling (CF) & 164 tubes \\
6 & Non-condensables in System (NC) & No nitrogen \\ \hline
\end{tabular}
\end{table}

To give the readers an intuition about the distribution of RP-1043 data, we employed \ac{LDA} to reduce the data into 2 dimensions, and visualized the dimension-reduced data in Fig.~\ref{fig:LDA}. As can be seen in the plots, FWE, FWC and NC faults are further away from normal than RL, RO and CL faults are. We can also see a general trend for a data point to deviate further away from the normal when the corresponding fault develops into a more severe level. It can also be seen from the plots that, SL1 data points are often closer to the fault-free region  than to their corresponding SL4 regions. In other words, these incipient fault data may look more like the fault-free data than their high-severity versions, which cast a challenge in using supervised learning algorithms (e.g., neural networks) to detect and diagnose these slight incipient faults.
\section{EXPERIMENTAL RESULTS}\label{sec:experiment}
\newcommand{\dprate}{0.1}

\subsection{Experimental Setup}
We conducted a case study to evaluate the performance of our MC-dropout approach on the RP-1043 dataset. We implemented the neural networks in Python using Keras~\cite{chollet2015keras}. The MC-dropout network $\mathcal{M}$ has four fully-connected hidden layers, each containing 20 nodes and with dropout layers interleaved. The output layer is a softmax layer with 7 output nodes, each representing an output class label (fault-free and the six faults in our study). Subscript $p$ indicates the dropout rate being used. By setting $p=0$, we will obtain a non-dropout network $\mathcal{M}_0$, which is used as a baseline to show the effect of MC-dropout on \ac{FDD} performance.

\subsection{Dropout Rate Selection}
As with many other machine learning models, the selection of hyperparameters of a MC-dropout network can have a significant impact on its prediction performance. In our case, we need to find a suitable dropout rate $p$, such that the resulting MC-dropout network can 1) accurately classify in-distribution examples with high confidence, and 2) identify ambiguous out-of-distribution examples by indicating high predictive uncertainty. The first requirement can be checked by using the usual cross-validation process; however, cross-validation will not help with the second requirement since we do not have access to the out-of-distribution data at training time.

\begin{figure*}[ptb]
    \centering
    \begin{subfigure}[b]{0.90\linewidth}
    \centering
    \includegraphics[width=\linewidth]{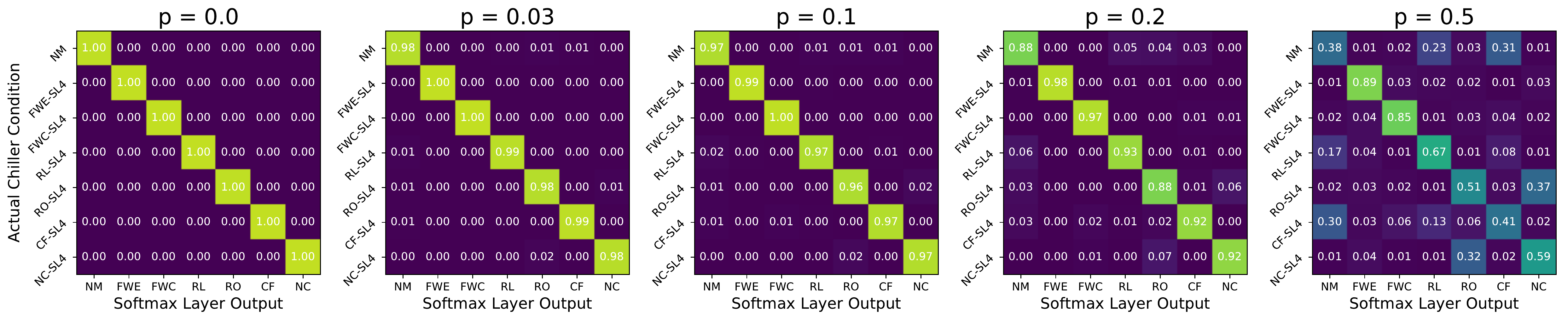}
    \caption{Predictive mean of MC-dropout network under different dropout rate settings}\label{fig:dp-rate-selection-mean}
    \end{subfigure}
    \\
    \begin{subfigure}[b]{0.90\linewidth}
    \centering
    \includegraphics[width=\linewidth]{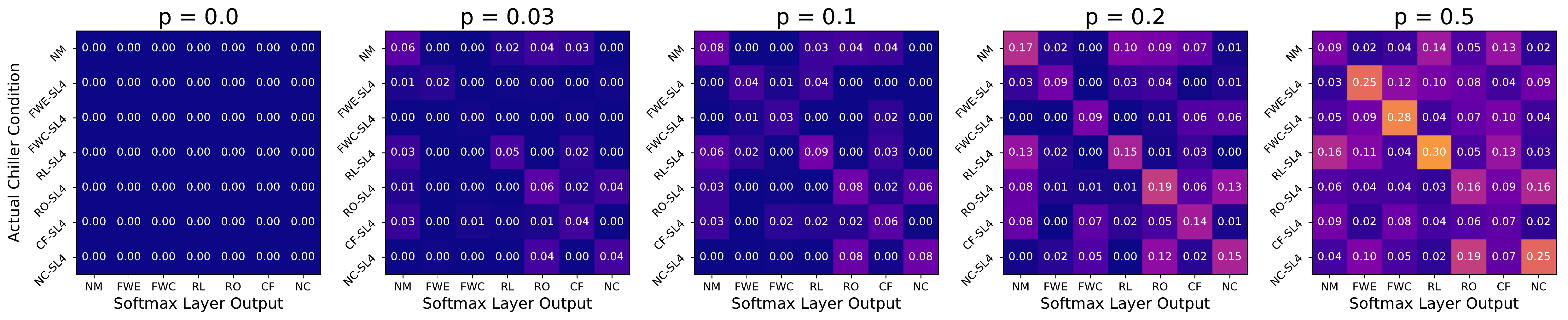}
    \caption{Predictive variance of MC-dropout network under different dropout rate settings}\label{fig:dp-rate-selection-var}
    \end{subfigure}
    \caption{The prediction results on in-distribution data, under five different dropout rate settings $p=0,0.03,0.1,0.2,0.5$. In the displayed heatmaps, each row corresponds to a chiller state (NM or a SL4 fault condition), and the values in each row are the respective predictive mean/variance at each output node.}\label{fig:dp-rate-selection}
\end{figure*}

In the experiment, we tested a number of dropout rates ranging from $0$ to $0.5$. Due to space limit, we only display in Fig.~\ref{fig:dp-rate-selection} the prediction results under five typical dropout rate settings ($p=0,0.03,0.1,0.2,0.5$). Each network was trained for 30 epochs, with categorical crossentropy used as the loss function. For each network with a non-zero dropout rate, $N=100$ Monte Carlo samples were drawn for estimating the predictive mean and variance.

With $p$ set to zero, our MC-dropout network will degenerate to a regular non-dropout neural network---no output variance will exist given the same input because the network does not have any inherent randomness. The network is thus not able to provide predictive uncertainty information.
When the dropout rate is high ($p=0.5$), the excessive dropout randomness as can be expected will undermine the model's predictive capability even on in-distribution examples as shown in Fig.~\ref{fig:dp-rate-selection}.  
In order to achieve a balance between the two aforementioned requirements, we used the following method to select a $p$ value empirically. We gradually increase $p$ until the predictive means (variances) start to drop (increase) fast. In this way, the trained model will still have good performance on in-distribution data, and can likely indicate out-of-distribution data via its predictive uncertainty. We chose $p=0.1$ for our subsequent experiments. Next, we are going to analyze the model’s performance on out-of-distribution data, and explain how the model’s outputs can be used to indicate potential faults.

\subsection{Comparison between Dropout and Non-dropout Methods}
\begin{figure*}[ptb]
    \centering
    \begin{subfigure}[b]{0.90\linewidth}
    \centering
    \includegraphics[width=\linewidth]{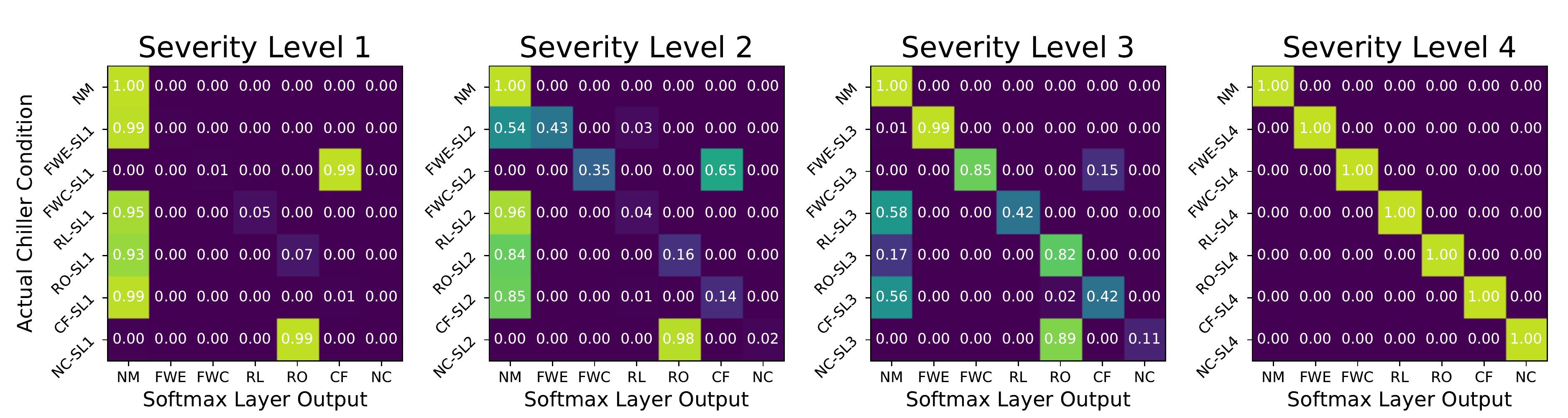}\vspace{-2mm}
    \caption{Softmax probability output of a non-dropout network ($p=0$)}
    \label{fig:mean-dp0}
    \end{subfigure}
    \\
    \begin{subfigure}[b]{0.90\linewidth}
    \centering
    \includegraphics[width=\linewidth]{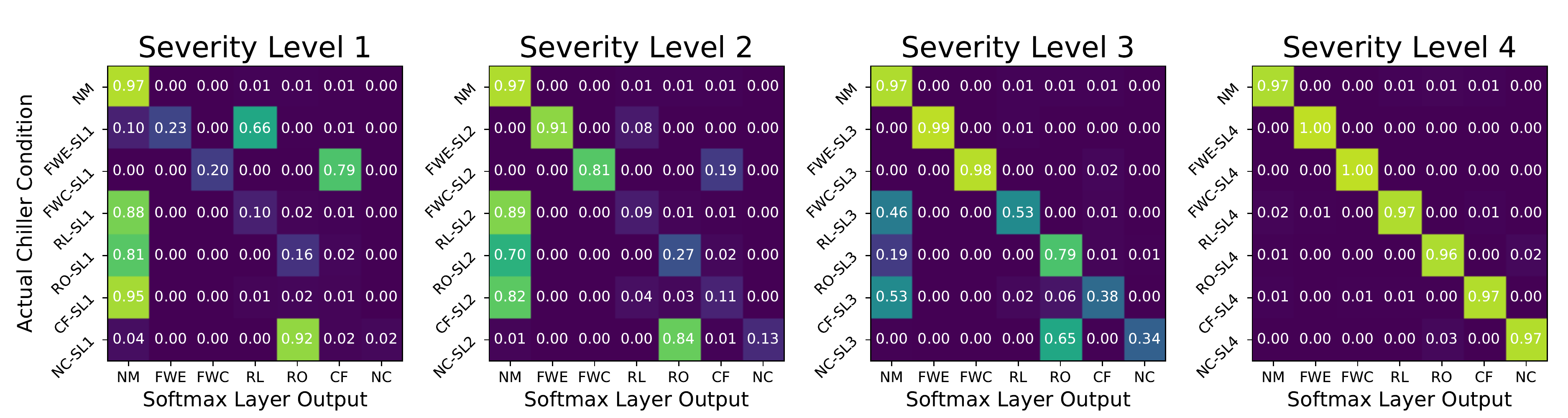}\vspace{-2mm}
    \caption{Predictive means of MC-dropout network ($p=0.10$)}
    \label{fig:mean-dp-med}
    \end{subfigure}
    \\
    \begin{subfigure}[b]{0.90\linewidth}
    \centering
    \includegraphics[width=\linewidth]{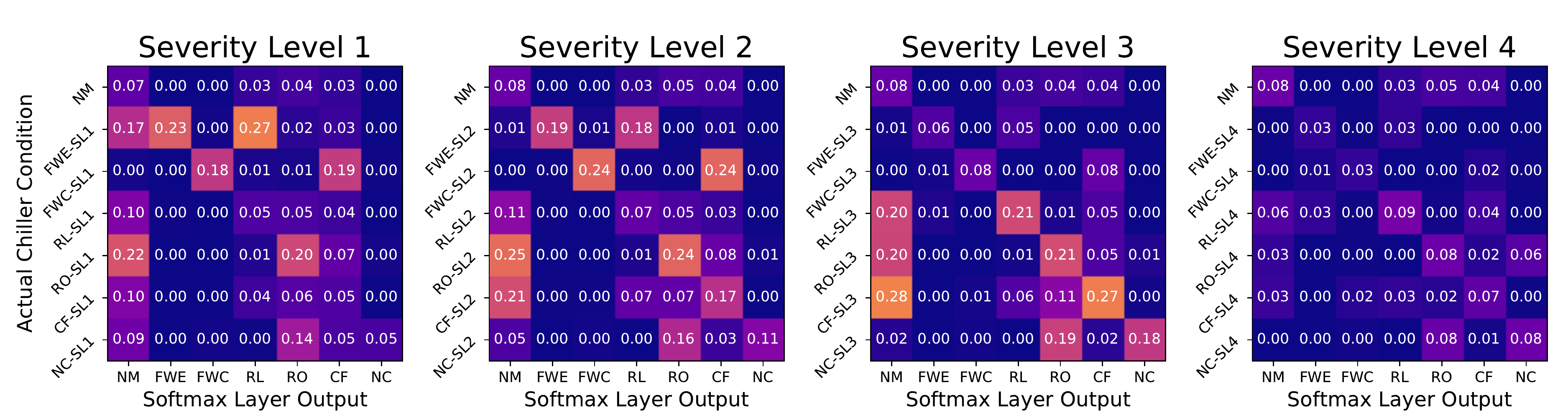}\vspace{-2mm}
    \caption{Predictive variances of MC-dropout network ($p=0.10$)}
    \label{fig:var-dp-med}
    \end{subfigure}
    \caption{Output comparison of the non-dropout network $\mathcal{M}_0$ and the MC-dropout network network $\mathcal{M}_{\dprate}$, on faults of all four severity levels.}\label{fig:heatmap}
\end{figure*}

We show in Fig.~\ref{fig:heatmap} a comparison between the non-dropout network $\mathcal{M}_0$ and the MC-dropout network $\mathcal{M}_{\dprate}$ with the chosen dropout rate $p=0.1$. Their prediction results on examples of all four severity levels are displayed. Again, for in-distribution (SL0 \& SL4) examples, both networks demonstrate good classification ability; in addition, the results given by $\mathcal{M}_{\dprate}$ shows little variance on in-distribution examples (SL0 \& SL4), which indicates its high confidence on these decisions.

SL3 faults do not appear in the training distribution, and are less severe than the SL4 faults used for training the models. As can be seen from Fig.~\ref{fig:heatmap}, both networks can still correctly recognize FWE-SL3, FWC-SL3 and RO-SL3 faults with high confidence; however, they also demonstrate some uncertainty about RL-SL3, CF-SL3 and NC-SL3 cases as can be seen from their predictive means. When the underlying fault is RL-SL3 or CF-SL3, the classifier is uncertain whether the chiller is in NM state, or the respective fault state. When the underlying fault is NC-SL3, the classifier is unsure about the underlying state being RO or NC. This phenomenon can be roughly explained using the \ac{LDA} analysis in Fig.~\ref{fig:LDA}. From the plots we can see that RL, RO, CF faults all reside very closely to the NM data points, which is one cause of the confusion. We can also see the presence of elevated predictive variance in blocks under suspicion; the predictive variance given by the MC-dropout network serves as another metric for indicating predictive uncertainty in such scenarios.

Our MC-dropout approach further demonstrates its usefulness in detecting and diagnosing slighter faults (SL1 \& SL2). It can be seen from the SL1 panel of Fig.~\ref{fig:mean-dp0} and Fig.~\ref{fig:mean-dp-med} that both classifiers make wrong predictions under all fault conditions, often with high confidence. SL1 fault conditions are classified as fault-free (as in RL-SL1, RO-SL1 and CF-SL1), or as other fault types (as in FWE-SL1, FWC-SL1 and NC-SL1). Similar problems are seen in SL2 cases as well. These phenomena can be partially understood using the visualization in Fig.~\ref{fig:LDA}. These low-severity fault data points often reside in the proximity of the data points of NM or other fault types, which results in the misclassification. Despite the ambiguity, the MC-dropout network $\mathcal{M}_{\dprate}$ casts its skepticism through the predictive uncertainty information given by Monte Carlo sampling. For example, when the chiller is under RO-SL1 fault, the predictive mean given by $\mathcal{M}_{\dprate}$ shows high confidence in believing the underlying state is NM, whereas the predictive variance shows high uncertainty in both NM and RO. The high predictive uncertainties indicate the possibility of a potential RO fault. 

We show the diagnostic results for all SL1 and SL2 fault conditions in Table~\ref{tbl:results}.
For the non-dropout network $\mathcal{M}_0$, we select class labels with softmax probability above $20\%$ as possible diagnoses. For the MC-dropout network $\mathcal{M}_{\dprate}$, we also include class labels with high predictive variance as possible diagnoses, in addition to those with softmax probability above $0.2$. A class label is considered to be with high predictive variance, if the ratio between the standard deviation of this particular class label and that of all class labels is above $10\%$. It can be seen from the table that $\mathcal{M}_0$ cannot correctly diagnose any of the SL1 faults, and is only able to correctly identify FWE-SL2 and FWC-SL2 among all SL2 faults. In comparison, the diagnoses given by the MC-dropout network $\mathcal{M}_{\dprate}$ contain the correct labels in all 12 cases listed in Table~\ref{tbl:results}. It can also be observed that $\mathcal{M}_{\dprate}$ is more certain about prediction results on SL2 cases than on SL1 cases. In FWE-SL1, RL-SL1, CF-SL1 and NC-SL1 cases, $\mathcal{M}_{\dprate}$ suggests three or more possible states, while in all SL2 cases the network only suggests two. This is understandable, because the fault signatures presented in SL1 cases are presumably less obvious than those in SL2 cases, thus creating less confusion for the neural network. Although the MC-dropout network is unable to give a definite diagnosis due to lack of information in such cases, it indicates some potential fault and narrows down the possible causes, which is valuable for further maintenance decisions to uncover the true status of the chiller system.
 
\begin{table}[tb]
\centering
\caption{Diagnosis Results}
\label{tbl:results}
\resizebox{\columnwidth}{!}{%
\begin{tabular}{lll}
\hline
Actual Chiller State & \begin{tabular}[c]{@{}l@{}}Diagnosis from \\ Non-dropout Network\end{tabular} & \begin{tabular}[c]{@{}l@{}}Diagnosis from \\ MC-dropout Network\end{tabular} \\ \hline
\textit{FWE-SL1} & NM & NM, \textbf{FWE}, RL \\
\textit{FWC-SL1} & CF & NM, \textbf{FWC}, CF \\
\textit{RL-SL1} & NM & NM, \textbf{RL}, RO, CF \\
\textit{RO-SL1} & NM & NM, \textbf{RO} \\
\textit{CF-SL1} & NM & NM, RL, RO, \textbf{CF} \\
\textit{NC-SL1} & RO & NM, RO, CF, \textbf{NC} \\
\textit{FWE-SL2} & NM, FWE & \textbf{FWE}, RL \\
\textit{FWC-SL2} & FWC, CF & \textbf{FWC}, CF \\
\textit{RL-SL2} & NM & NM, \textbf{RL} \\
\textit{RO-SL2} & NM & NM, \textbf{RO} \\
\textit{CF-SL2} & NM & NM, \textbf{CF} \\
\textit{NC-SL2} & RO & RO, \textbf{NC} \\ \hline
\end{tabular}%
}
\end{table} 
 
From the above analyses, we can see that the uncertainty information given by MC-dropout networks is useful in cases where there is a lack of information about the characteristics of incipient faults. Under such situations, it is better and more reasonable for a classifier to show the uncertaintiy about its decisions, rather than just giving definite but often incorrect predictions. We can also see that the softmax probabilities alone have limited capability in telling the prediction uncertainty of a neural network, which limits its use in detecting unseen, incipient fault conditions. With MC-dropout, the network has gained another way to indicate its uncertainty through the predictive variance. Our experimental results on RP-1043 have demonstrated its usefulness in detecting and diagnosing incipient faults.


\section{CONCLUSIONS}\label{sec:conclusion}
In this paper, we proposed using MC-dropout, a method for estimating a deep learning model's uncertainty in its decisions, to detect incipient or unknown faults in modern \acp{CPS} with a neural network trained with limited fault data. By presenting a case study on ASHRAE RP-1043 dataset, we have shown the effectiveness of MC-dropout in detecting and diagnosing chiller faults. As part of our future work, we plan to conduct a more theoretical analysis on the proposed approach to gain a better understanding of it.

\section*{ACKNOWLEDGMENT}
This work is supported in part by the National Research Foundation of Singapore through a grant to the Berkeley Education Alliance for Research in Singapore (BEARS) for the Singapore-Berkeley Building Efficiency and Sustainability in the Tropics (SinBerBEST) program, and by the National Science Foundation under Grant No.~1645964. BEARS has been established by the University of California, Berkeley as a center for intellectual excellence in research and education in Singapore.

\bibliographystyle{plain}
\bibliography{refs}

\end{document}